\documentclass[11pt, letterpaper]{article}

\usepackage[margin=1in]{geometry} 
\usepackage{amsmath}              
\usepackage{graphicx}             
\usepackage{hyperref}             
\usepackage{url}                  
\usepackage{float}

\usepackage{tgpagella,eulervm}

\usepackage{caption}
\usepackage[activate={true,nocompatibility},final,tracking=true,kerning=true,spacing=true,factor=1100,stretch=10,shrink=10]{microtype}

\hypersetup{
    colorlinks=true,
    linkcolor=blue,
    filecolor=magenta,      
    urlcolor=cyan,
    citecolor=red,
    pdftitle={From Taylor Series to Fourier Synthesis: The Periodic Linear Unit},
    pdfauthor={Shiko Kudo},
}

\title{From Taylor Series to Fourier Synthesis: \\ The Periodic Linear Unit}
\author{Shiko Kudo \\ National Taiwan University}
\date{August 1, 2025}

\begin{document}

\maketitle

\begin{abstract}
The dominant paradigm in modern neural networks relies on simple, monotonically-increasing activation functions like ReLU. While effective, this paradigm necessitates large, massively-parameterized models to approximate complex functions. In this paper, we introduce the Periodic Linear Unit (PLU), a learnable sine-wave based activation with periodic non-monotonicity. PLU is designed for maximum expressive power and numerical stability, achieved through its formulation and a paired innovation we term Repulsive Reparameterization, which prevents the activation from collapsing into a non-expressive linear function. We demonstrate that a minimal MLP with only two PLU neurons can solve the spiral classification task, a feat impossible for equivalent networks using standard activations. This suggests a paradigm shift from networks as piecewise Taylor-like approximators to powerful Fourier-like function synthesizers, achieving exponential gains in parameter efficiency by placing intelligence in the neuron itself.
\end{abstract}

\section{Introduction}

The dominant paradigm in modern neural networks relies on simple, monotonically-increasing or generally monotonically-increasing activation functions like the Rectified Linear Unit (ReLU). While effective, these activations necessitate large, massively-parameterized models, as complex functions must be approximated by combining a vast number of simple, piecewise linear components.

Recent work has explored more complex activations. For instance, GELU \cite{hendrycks2016gaussianerrorlinearunits} offers a smooth approximation of ReLU, and while technically non-monotonic, its behavior is dominated by a monotonic increase. Periodic activations like Snake \cite{ziyin2020neuralnetworksfaillearn} imbue networks with a more suitable inductive bias for periodic data, but they are formulated as strictly non-decreasing functions as well. These functions enrich the network's capabilities not through a fundamental change in the paradigm of using neurons, but by utilizing and modifying the standard generally monotonically-increasing curved activation function paradigm that has defined the last decade of Machine Learning research.

In this paper, we introduce the Periodic Linear Unit (PLU), a novel, learnable sine-wave based activation function with periodic non-monotonicity by formulation, designed to provide maximum expressive power while ensuring numerical stability. We achieve this through careful formulation, as well as a paired key innovation we term Repulsive Reparameterization, a mechanism that architecturally prevents the novel activation from collapsing into a non-expressive linear function during training.

We demonstrate on the spiral classification task that this design leads to a fundamental shift in network capability. A minimal MLP with only two PLU neurons can successfully solve the problem, a task impossible for networks of an equivalent size using ReLU, GELU \cite{hendrycks2016gaussianerrorlinearunits}, or Snake \cite{ziyin2020neuralnetworksfaillearn}. This suggests a paradigm shift from networks as piecewise approximators to powerful function synthesizers, achieving exponential gains in parameter efficiency by placing intelligence in the neuron itself.

\section{Early Formulation}

Before we reach the final formulation, let us begin by examining a simpler formulation, and set the scene for where the formulation originated from initially.

The problem begins with audio synthesis, where we would like to introduce an inductive bias for repeating signals. However, while Snake is one such formulation that is already widely adopted, it had certain restrictions, such as not being an odd function. Could we achieve similar results using a simple, odd, pure sine wave? Let us put aside the implications of what we would be turning the neural network into and save it for later, focusing only on this idea for now. Our theoretical first draft for the PLU is as follows:
\begin{equation}
    \text{PLU}_{\text{theoretical}}(x, \alpha, \beta) = x + \frac{\beta}{1 + \lvert\beta\rvert} \sin(\lvert\alpha\rvert x)
    \label{eq:plu_theoretical}
\end{equation}
where $\alpha$ and $\beta$ are learnable parameters controlling the frequency and amplitude.

The $x$ component serves as a residual path. By including it, the function is identity-like at its core, allowing gradients to flow through it easily just like a residual connection. It prevents the vanishing gradient problem and makes training much more stable.

The $\sin(x)$ component then provides periodic non-linearity.

\subsection{Choices Made}

The scaling factors are important hyperparameters. $\beta$ provides balancing between the sine wave and the linear component. $\beta=1.0$ is a good default value, as it strikes a delicate balance between the periodic gradient "deadzone" that begins occurring every $\pi$ cycles if it is set much higher, and the completely non-periodic simple linear unit obtained from setting it to $0.0$. $\alpha$ provides the means for modifying the period of the activation.

There are some important reasons why $\alpha$ and $\beta$ are separate parameters within this activation. Specifically, the $\frac{\beta}{1 + \lvert\beta\rvert}$ term provides several tangible benefits:
\begin{enumerate}
    \item While the model is given freedom to increase the oscillation of the activation through changing the $\beta$, there is a point at which such oscillations become a major issue. A rapidly oscillating activation deteriorates the loss landscape significantly with its severe non-monotonicity, and can lead to numerical stability issues or training runs that have to be stopped midway. Having the scaling term be formulated as $\frac{\beta}{1 + \lvert\beta\rvert}$ however, means that the activation discourages runaway activations like this. It discourages the model from introducing wildly fluctuating activations, as you have to have $\beta$ become very large to obtain such results. The maximum oscillation which can be introduced by the model is $x + \sin(x)$ at $\beta \to \infty$, effectively capping the oscillating behavior at a well-behaved range.
    \item The activation can provide a very good selection of shapes with $\alpha$ and $\beta$ both being just around the $-3.0 < \text{value} < 3.0$ range, which is very good considering optimizers tend to prefer weights in that general range as well. You can get a flipped phase activation by flipping the sign on $\beta$, you can get non-monotonicity with the right combinations, and more.
    \item An activation in the same close family of PLU is Snake, which is formulated as $x + \frac{1}{a} \sin^2(ax)$. As we will notice immediately, if the learnable parameter $a$ needs to cross zero during training (e.g., to flip the "phase" of the non-linearity), the model encounters a $1.0/0.0$ singularity. This results in NaNs and a crashed training run, and is usually fixed by a custom autograd function. PLU however does not require this: Since $\lvert\beta\rvert$ is always non-negative ($\ge 0$), the denominator $1 + \lvert\beta\rvert$ is always greater than or equal to 1. Hence, there is no possibility of division by zero. The function is perfectly defined and stable for all possible real values of $\beta$.
\end{enumerate}

Another distinct feature is the $\lvert\alpha\rvert$ term. Why is there an absolute value function applied to $\alpha$? This was chosen based on the consideration of phase-flipping the activation function. One of the features of PLU is the ability for the model to flip the phase of the oscillation if necessary, but a crucial consideration when giving the model such an ability is then the various avenues with which it can achieve this. Importantly, notice that without the absolute value on the $\alpha$, the model has two ways of flipping the phase: either by letting $\beta$, the periodicity strength parameter cross $0.0$, or by having $\alpha$ cross zero. Imagine that a specific gradient step pushes the model in the direction of flipping the phase. What would happen to the gradients at both those parameters? The gradient step might push both $\alpha$ \textit{and} $\beta$ across $0.0$. But if this step is taken, then the phase flip is essentially cancelled. We flip it twice and end up with the same phase. This is not good for the loss landscape, and so we must choose one of these mechanisms and stick to using it for flipping the oscillatory behavior. It is then much more natural to use $\beta$ for this purpose, as crossing $0.0$ with $\beta$ does not introduce any wild changes with the activation, while $\alpha$ crossing $0.0$ pushes the scale of the entire activation horizontally and massively, letting the period of the oscillation grow to infinity, before coming back to a reasonable range once again with flipped phase.

\subsection{The Untapped Potential of Non-Monotonicity}

At certain values of $\alpha$ and $\beta$, namely when $1 + \frac{\beta}{1 + \lvert\beta\rvert} \lvert\alpha\rvert \cos(\lvert\alpha\rvert x) < 0$ for some $x$, PLU turns into a non-monotonically increasing function. This is not the case for the default $(\alpha, \beta)$ values of $(1.0, 1.0)$, but in a parametric setting, it is possible for the model to learn this behavior.

To control the amount of non-monotonicity allowed, as we have discussed already in the point above about capping the oscillating behavior to a well-behaved range, this is done by making non-monotonicity of an excessive degree harder to obtain by the model, as a large $\beta$ of around $10.0$ is necessary to even approach a $0.9$ influence from the sine wave relative to the residual $x$. Optimizers tend to prefer moderately sized parameters, and SGD steps are small, meaning this is a subtle push that encourages the model to utilize moderate amounts of the periodic component and focus on improving the model elsewhere.

However, the non-monotonicity is by no means a bug. It is one of the central features of PLU.

While it is hard to gauge the impact of such periodic non-monotonicity during model training, one hypothesis as to a situation that might happen is that unlike with monotonic activations like ReLU or Snake, the model might learn to "settle" in a specific range of the activation. This is optimal compared to the current standards of activations, since a repeating non-monotonic activation provides a chance to easily "escape" any zero gradient zones with a good optimizer, while activations such as Sigmoid and ReLU both have a near zero gradient for most of its domain and half of its domain respectively. PLU effectively provides an infinite amount of these "window of complete activation functions", these repeating complex gradient regions, that the model is able to access and potentially settle into based on its needs, and then a small nudge to the input $x$ is all that is needed to push it into a different region where the gradients are large again, if it settles within a low-gradient region. Zero-gradient spots only occur at incredibly sparse intervals on points, rather than for entire ranges of values in the domain. This can lead to more dynamic and effective training by balancing the problem of the vanishing gradient against not having any such gradient-based regularization at all, as is the case with many monotonic activations. There is also a more fundamental difference, however, which we will discuss now.

By taking a closer look at the potential differences, we can quickly ascertain that compared to monotonic activations like ReLU when $x>0$ or Snake, the training dynamics will be entirely different, as in those cases, the monotonic activations basically act as transparent layers that don't affect the SGD steps per-se. At each step, monotonic activations don't fundamentally alter the direction of the gradient since $dA/dx$ (where $A$ is the activation) is always positive and by the chain rule, applying said partial derivative will never flip the sign; thus, these activations end up acting more like a dynamic learning rate than a complex function shaper.

With this in mind then, there are many theoretical benefits in using a non-monotonic activation instead:
\begin{itemize}
    \item \textbf{Increased Representational Power:} A non-monotonic activation can approximate more complex functions with fewer parameters. A single neuron can now model a "bump" or a "wiggle" rather than just a simple "on/off" or "saturation" behavior. For a network to model a bump with monotonic activations, it would need at least two values working in opposition (e.g., $\text{sigmoid}(x) - \text{sigmoid}(x-c)$). This implicitly puts more emphasis on training cross-neuron relations across a massive number of simple neurons, which becomes much less crucial when a single neuron exhibits significantly more capabilities.
    \item \textbf{Learning "Valleys" and "Bands":} This is one already mentioned, but it is worth going into it slightly more in depth. The model might learn to "settle" its pre-activations into a specific range. This is a form of implicit regularization or feature selection. The network could learn that for a certain type of input, the optimal feature value lies in one of the "valleys" of the activation function. This is a much richer behavior than simply pushing a value as high as possible, as you would with monotonically increasing activations. It allows the network to learn not just "more or less is better for this activation" but "this specific range is better."
\end{itemize}

One of the last key things to note about PLU then, though, is that it is only non-monotonic for certain choices of parameters, as stated at the very start of this section. This is good, as it means the model has a choice as to whether it would like to make use of the non-monotonicity or not. If it finds that it prefers a simple monotonic activation for a specific layer, it can do so, at which point PLU reverts to providing the more standard transparent activation represented by functions such as ReLU.

\subsection{Asymptotic Behavior}
\begin{itemize}
    \item If $\beta = 0$, the $\frac{\beta}{1 + \lvert\beta\rvert}$ term is 0. (The activation is the identity).
    \item As $\beta \to +\infty$, the term approaches $+1$, and the activation becomes $x + \sin(\alpha x)$.
    \item As $\beta \to -\infty$, the term approaches $-1$, and the activation becomes $x - \sin(\alpha x)$.
    \item If $\beta = 1$, the term is $1/2 = 0.5$. The activation becomes $x + 0.5 \sin(\alpha x)$, where the derivative at $x=0$ is $1.5$.
    \item If $\beta = -1$, the term is $-1/2 = -0.5$. The activation becomes $x - 0.5 \sin(\alpha x)$, where the derivative at $x=0$ is $0.5$.
    \item As $\alpha \to 0$, the activation approaches the identity.
\end{itemize}

\section{The Collapse to Linearity in Practice}

The early formulation which we have covered in depth, appears to be valid in design. However, how does it perform in practice?

While the theory up until this point is fairly sound, if we were to run the spiral example now without the reparameterization/regularization terms, we will be met with disappointing results. ReLU, GELU, and Snake will all attempt to begin optimizing in the 8 neurons example as well as the 2, but PLU without regularization terms appears to be stagnant from the first step. What could be the issue?

The answer lies within training dynamics. One of the points of consideration during our Early Formulation was the elimination of regions in the activation which were undefined. This is valid, and with a fixed $\alpha$ and $\beta$, PLU without parametrization becomes a working, regular activation. However, the power of PLU comes not from fixing its parameters but in allowing the optimizer to find the best parameters for it as well.

Yet the moment we introduce such learning of parameters, the learning halts, because the optimizer is essentially faced with two paths at the beginning of training:
\begin{enumerate}
    \item Attempt to navigate the complicated path of learning to utilize the PLU as an oscillating activation function, with $\alpha \neq 0.0$ and $\beta \neq 0.0$.
    \item Or two, simply push both $\alpha$ and $\beta$ to zeros immediately and obtain a simple linear function.
\end{enumerate}
The optimizer almost always picks the latter approach. Since PLU can be non-monotonic, something unusual for activations, the model quickly finds that pushing the parameters of the PLU to zero and obtaining a "safe average" activation of a linear line without such non-monotonicity is best based on its immediate gradients on the chaotic loss plane; and once the model pushes $\alpha$ and $\beta$ into that region, we find that training halts entirely.

This means that perhaps, having the entire range of parameters in PLU be 'safe' was not desirable after all.

\section{Repulsive Reparameterization}

The key in solving the problem of helping the optimizer learn to utilize the complex activation function is the often reinforced observation about the path of least resistance. If there are two directions which the model weights can go in, one that immediately raises the loss immensely but with the unknowable potential of a lower loss beyond the cliff, and another that promises to lower the loss slightly, the gradient step will overwhelmingly be in the direction of lowering the loss. Mechanisms such as the RMSProp \cite{hintonetalrmsprop} or Adam \cite{kingma2017adammethodstochasticoptimization} optimizers attempt to fix this, but even they will not let a model take a step that appears to increase the loss ten-fold.

Thus, the key in forcing the optimizer to use PLU to its full effect turns out to be in purposefully making the increase in loss from turning it into a linear function immense. While previously, the optimizer might have, at the very first step, seen the two paths and decided that the loss can be lowered by reducing the PLU to a linear function, we now seek for the optimizer to instead see a massive towering cliff in front of the path towards a more linear PLU, while the path away is unimpeded.

Specifically, the problematic values for both $\alpha$ and $\beta$ are when they are $0.0$. So, in effect we hope that the penalty in the gradients for lowering loss grows more and more as the optimizer pushes PLU towards having near-zero $\alpha$ and $\beta$'s, thus discouraging the optimizer from collapsing the activation into a linear identity.

Asymptotic behavior from $1 / x$ fits this objective perfectly.

We now introduce the final formulation of PLU.

\section{The Periodic Linear Unit}

A \textbf{Periodic Linear Unit (PLU)} is composed of a scaled linear sum of the sine function and $x$.
\begin{equation}
    \text{PLU}(x, \alpha, \rho_\alpha, \beta, \rho_\beta) = x + \frac{\beta_{\text{eff}}}{1 + \lvert\beta_{\text{eff}}\rvert} \sin(\lvert\alpha_{\text{eff}}\rvert x)
    \label{eq:plu_final}
\end{equation}
Where the effective parameters, $\alpha_{\text{eff}}$ and $\beta_{\text{eff}}$, are reparameterized using learnable repulsion terms, $\rho_\alpha$ and $\rho_\beta$ as follows:
\begin{align}
    \alpha_{\text{eff}} &= \alpha + \frac{\rho_\alpha}{\alpha} \\
    \beta_{\text{eff}} &= \beta + \frac{\rho_\beta}{\beta}
\end{align}

As the optimizer attempts to push $\alpha$ towards $0.0$, the $\rho_\alpha / \alpha$ term grows quickly, exploding to infinity and thus making the period of the sine component become small at an alarming rate. While many real world data points have the tendency to have repeating patterns in them, there is none that will benefit in any way from a sine component with a period that is quickly growing infinitely small. Thus, the optimizer is actively pushed back against its desires to push $\alpha$ towards $0.0$ by an explosion of loss when taking even a small step in this direction.

As the optimizer attempts to push $\beta$ towards $0.0$, the $\rho_\beta / \beta$ term causes $\beta_{\text{eff}}$ to explode towards positive or negative infinity. Consequently, the amplitude scaling term $\beta_{\text{eff}} / (1 + \lvert\beta_{\text{eff}}\rvert)$ rapidly approaches its maximum absolute value of 1. The periodic component becomes "stuck" at full amplitude ($\pm\sin(\lvert\alpha_{\text{eff}}\rvert*x)$), and the optimizer thus loses all ability to reduce its influence. To regain control and continue minimizing the loss, it is then forced to push $\beta$ away from the zero-crossing, thus avoiding the linear identity function.

The most crucial part of this reparameterization is that for a sufficiently large $\alpha$ and $\beta$, the reparameterization does not occur. The original $\alpha$ and $\beta$ are simply passed through, as both $\rho_p / p$ terms approach $0.0$ as $p=(\alpha \text{ or } \beta)$ grows. This is the mathematical formulation of our "gradient wall" idea. In the end, all that is affected is the asymptotic behavior at $\alpha=0.0$ and $\beta=0.0$, and the full expressive behavior of PLU is left fully intact. We term this mechanism \textbf{Repulsive Reparameterization}.

\subsection{Choice of \texorpdfstring{$\rho_\alpha$ and $\rho_\beta$}{rho\_alpha and rho\_beta}}

While these parameters are best learned by the optimizer itself, the initial guess provided can have an important effect. In picking this starting value, it is useful to further examine the reparameterization equation:
\begin{equation}
    R(P, \rho_P) = P + \frac{\rho_P}{P}
    \label{eq:reparam}
\end{equation}
where $P = [\alpha, \beta]$ and $\rho_P = [\rho_\alpha, \rho_\beta]$.
The derivative with respect to $P$ is:
\begin{equation}
    \frac{dR}{dP} = 1 - \rho_P P^{-2}
\end{equation}
Let us say $P_0$ is the point at which the above derivative is zero. Then, we obtain:
\begin{align*}
    \rho_P P_0^{-2} &= 1 \\
    P_0^{-2} &= \frac{1}{\rho_P} \\
    P_0 &= \sqrt{\rho_P}
\end{align*}
Thus, the reparametrization scheme in effect limits the effective range of $P$ to $P_{\text{eff}} \ge 2 P_0$, or more intuitively $P_{\text{eff}} \ge 2 \sqrt{\rho_P}$.

Putting this in context with the two parameters of PLU:

$\alpha$ is the period multiplier, and a larger value produces a PLU which oscillates faster. A lower value on the other hand creates longer oscillations. In this case then we should set $\rho_\alpha$ based on the lower bound of how wide of an activation oscillation we want to allow. A $\rho_\alpha$ of $5.0$, as an example, will allow any $\alpha_{\text{eff}} \ge 2 \sqrt{5.0} \approx 4.472$. An overly large $\rho_\alpha$ will force the model into utilizing a PLU activation with a fast oscillation by enforcing a minimum frequency on the sin component in this way, the effects of which are explored in our 'High-Frequency Prior' experiment (Section \ref{sec:chaotic_init}, ``The 'Chaotic Initialization' Paradigm''). The effects of this are interesting to say the least, and will be explored later, as they indicate a profound shift in how neural networks can learn.

$\beta$ is then the soft-bounded strength of the sine component. In the previous section it was mentioned that a $\beta$ value of $0.0$ corresponds to a linear function, and as it is increased, slowly, more of the sine component is introduced. $\rho_\beta$ then provides a lower bound for $\beta$, forcing the model to never be able to go below a specific sine wave contribution fraction. For example, a $\rho_\beta$ of $0.15$ will allow only $\beta_{\text{eff}} \ge 2 \sqrt{0.15} \approx 0.775$, or in direct terms a minimum multiplier on the sin component of $\beta_{\text{eff}} / (1 + \lvert\beta_{\text{eff}}\rvert) \approx 0.437$.

\section{Experiments and Results}

To validate the efficacy and unique properties of the Periodic Linear Unit (PLU), we conduct a series of experiments on the classic spiral classification task. This task is notoriously difficult for simple statistical models as it requires learning a highly non-linear, curved decision boundary. We compare PLU against three standard activation functions: ReLU, GELU, and the periodic Snake activation.

\textit{\textbf{Experimental Setup:}} All models are simple Multi-Layer Perceptrons (MLPs) trained with the Adam optimizer (lr=0.01) and Binary Cross-Entropy loss. A crucial constraint is applied to all experiments: for both PLU and Snake which are parametric, the learnable activation parameters are \textbf{shared} across all neurons in the entire network. This provides a challenging test case, as the network must find a single activation shape which is suitable for all neurons.

\subsection{Superior Convergence and Expressivity in a Standard MLP}

Our first experiment utilizes a standard MLP architecture with two hidden layers of 8 neurons each (2-8-8-1). We initialize PLU with moderate repulsion ($\text{init\_rho\_alpha}=5.0$, $\text{init\_rho\_beta}=0.15$) (The Adam optimizer begins with these initial values of $\rho_\alpha$ and $\rho_\beta$ and is allowed to modify them).

\begin{figure}[h!]
    \centering
    \includegraphics[width=0.32\textwidth]{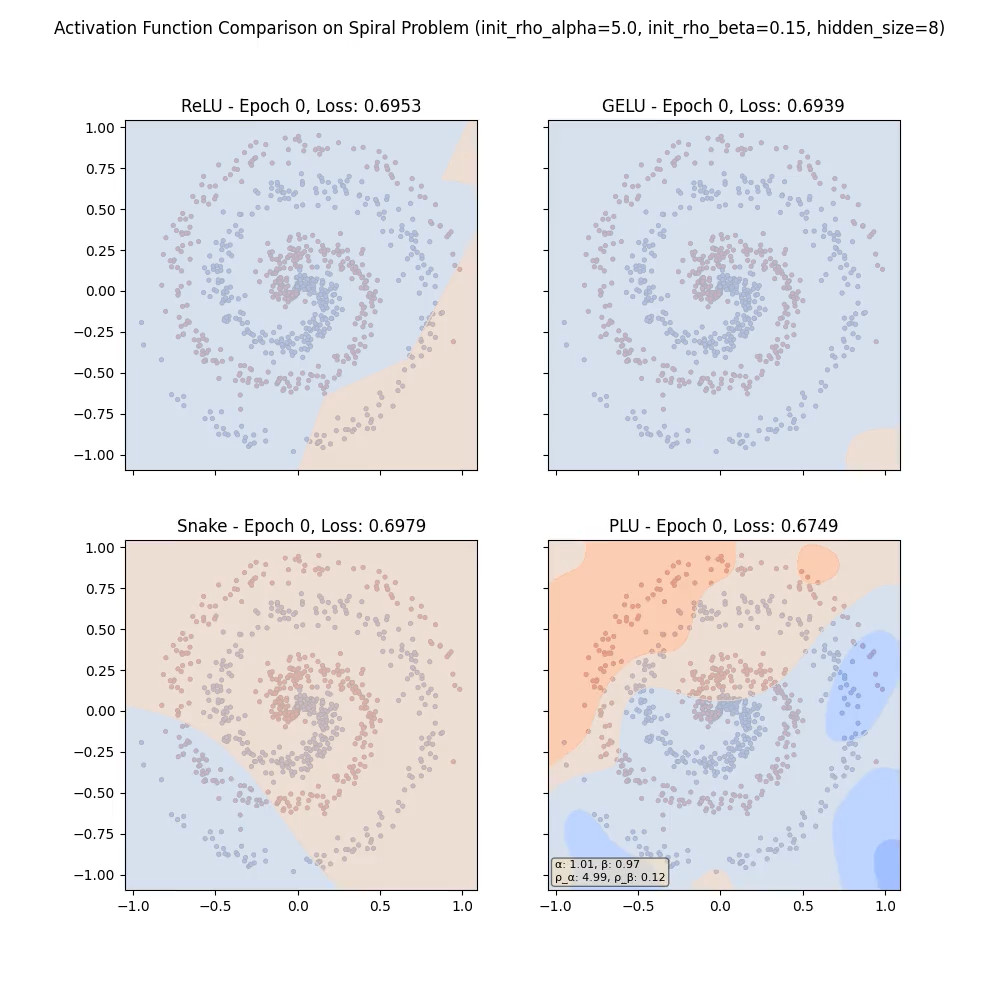}
    \includegraphics[width=0.32\textwidth]{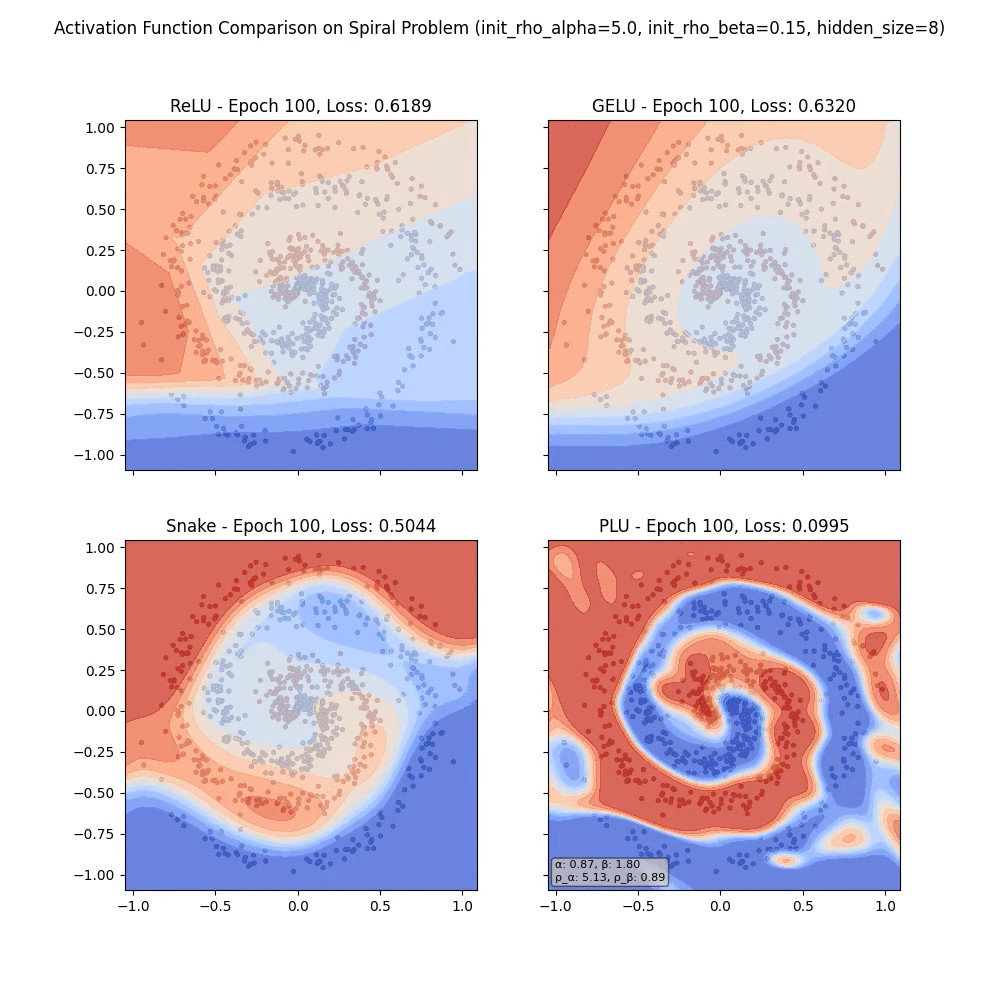}
    \includegraphics[width=0.32\textwidth]{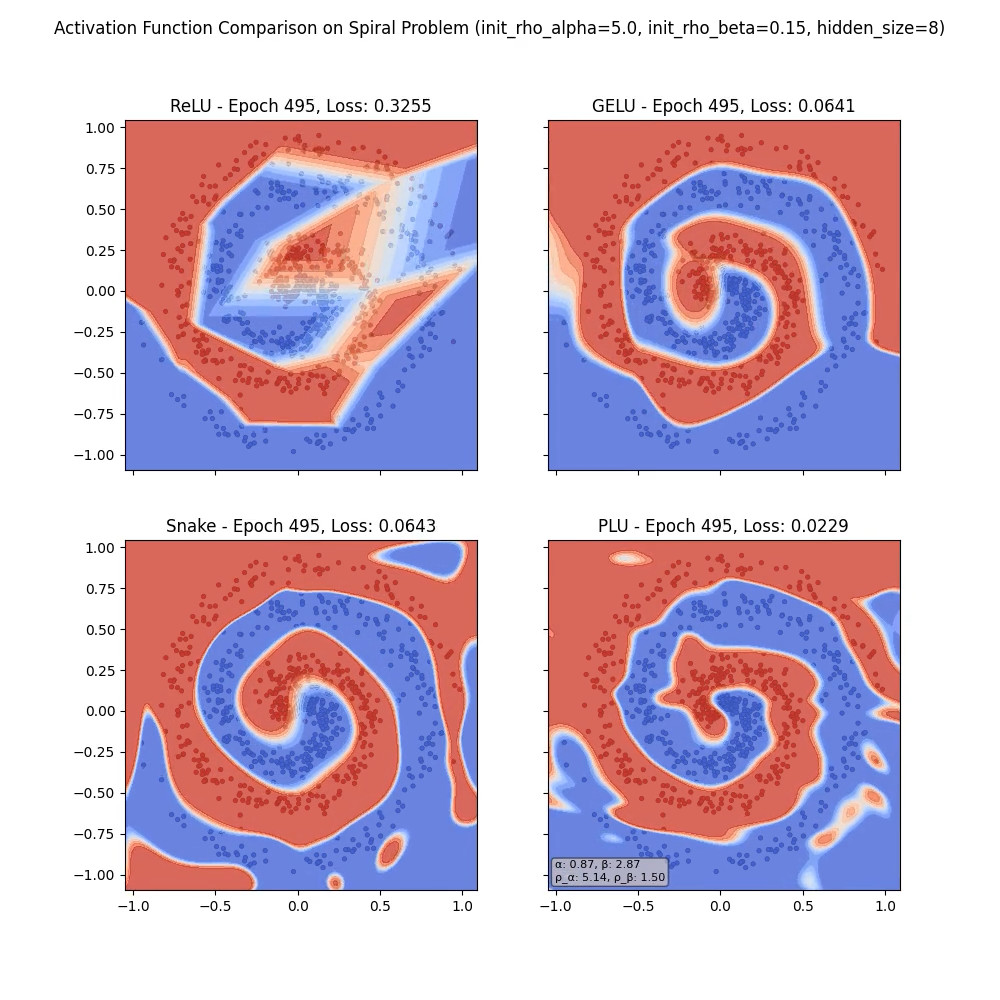}
    \caption{Time-series comparison of decision boundary learning with an 8-neuron MLP. (a) At Epoch 0, all models start with simple, large-scale boundaries. (b) By Epoch 100, PLU has nearly solved the problem while others are still forming the basic shape. (c) At Epoch 495, PLU has achieved the lowest loss and a highly accurate boundary.}
    \label{fig:spiral_8_neurons}
\end{figure}

As shown in Figure \ref{fig:spiral_8_neurons}, at only 100 epochs, the PLU-based network achieves a loss of \textbf{0.0995}, effectively solving the classification task. In contrast, the next best competitor, Snake, is at 0.5044, with ReLU and GELU lagging significantly. By the end of the run (Epoch 495), PLU maintains the lowest loss at \textbf{0.0229}.

There is something else that is interesting however that we already begin to see with this example. The results demonstrate a dramatic difference in learning dynamics. ReLU, GELU, and Snake appear to learn by progressively "curving" or "folding" an initially simple decision boundary to fit the data.

PLU, however, seems to form a more complex initial structure that rapidly "emerges" or "coalesces" into the correct spiral shape. We will see more of this with the following examples.

\subsection{The ``Chaotic Initialization'' Paradigm}
\label{sec:chaotic_init}

To explore the effect of the repulsive reparameterization, we conduct a second experiment with the same 8-neuron 2-8-8-1 MLP but initialize PLU with extreme repulsion terms ($\text{init\_rho\_alpha}=15.0$, $\text{init\_rho\_beta}=0.5$). This forces the effective $\alpha$ and $\beta$ to be large from the outset, inducing high-frequency oscillations. Specifically, the earlier configurations for $\rho$ sets a lower-bound for the optimizer on $\alpha$ and $\beta$ of $\alpha_{\text{eff}} \ge 4.472$ and $\beta_{\text{eff}} \ge 0.775$, while with the new configurations, this lower floor is lifted all the way to $\alpha_{\text{eff}} \ge 7.746$ and $\beta_{\text{eff}} \ge 1.414$. The effect is that the contribution from the sine wave is forced to be very high throughout the run, and the oscillation period of that sine wave is also kept incredibly short.

\begin{figure}[H]
    \centering
    \includegraphics[width=0.8\textwidth]{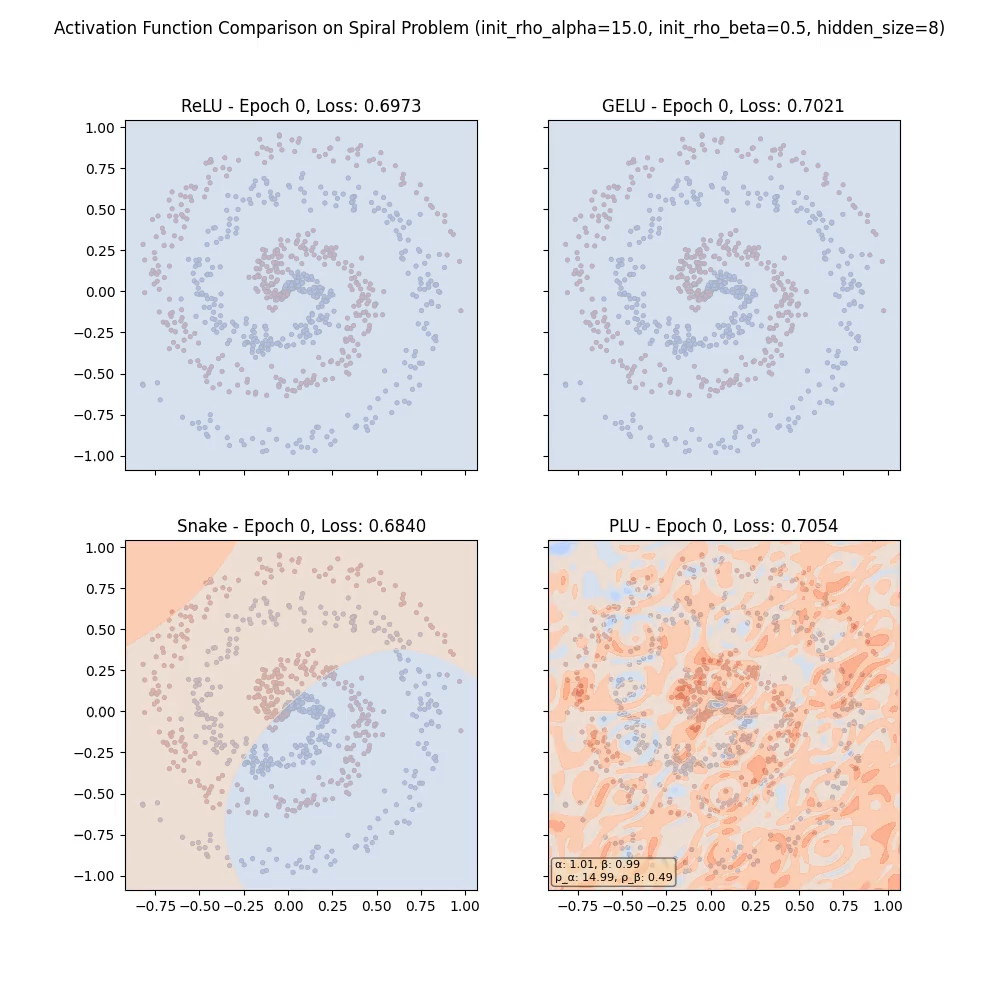}
    \caption{The effect of a high-frequency prior at Epoch 0. While other activations initialize with smooth boundaries, PLU begins with a complex, high-frequency "chaotic" state.}
    \label{fig:chaotic_init}
\end{figure}

\begin{figure}[H]
    \centering
    \includegraphics[width=0.8\textwidth]{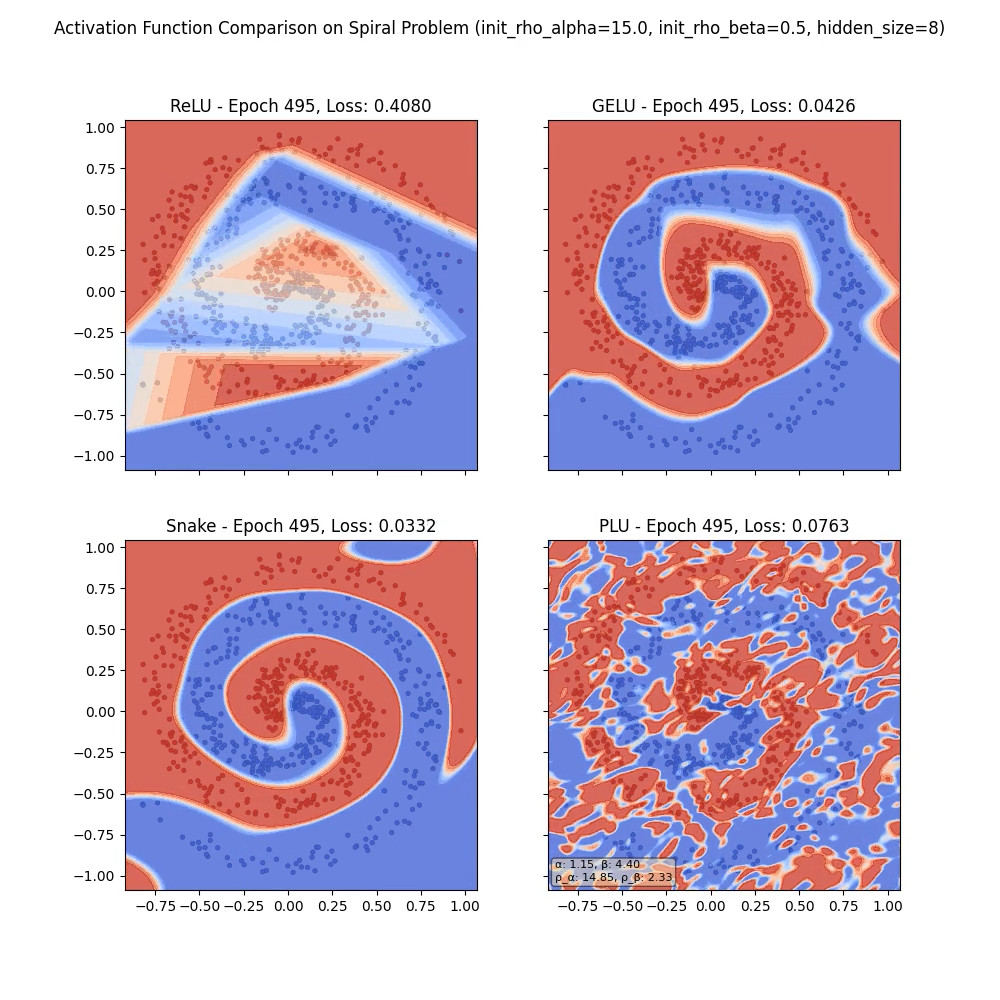}
    \caption{Final state at Epoch 495 for the high-frequency prior experiment. While Snake and GELU solve the task, the PLU network struggles to adjust its complex initial state, resulting in a higher final loss, but the shape of the final decision boundaries are nonetheless visibly the spiral.}
    \label{fig:chaotic_final}
\end{figure}

This experiment is not designed to achieve the lowest loss, but to reveal the underlying mechanism of PLU. As seen in Figure \ref{fig:chaotic_init}, the PLU network does not start with a simple function, but is instead \textbf{initialized with immense complexity}. The decision boundary at Epoch 0 is already a high-frequency, chaotic energy landscape.

This reveals a completely different learning paradigm. Instead of starting simple and learning to be complex, the PLU network appears to \textbf{start complex and learn to simplify}. The optimizer's task becomes one of "taming" the oscillations to fit the data, rather than shaping a simpler structure into form. While this strong prior ultimately hinders convergence on this specific task (Figure \ref{fig:chaotic_final}), it demonstrates that PLU's expressivity can be controlled at initialization to a degree impossible with other activations.

\subsection{Solving the Spiral with Two Neurons}

The final and most definitive experiment reduces the network to its absolute minimum: an MLP with two hidden layers of only 2 neurons each (2-2-2-1). This task is theoretically impossible for a ReLU-based network of this size.

\begin{figure}[H]
    \centering
    \includegraphics[width=0.8\textwidth]{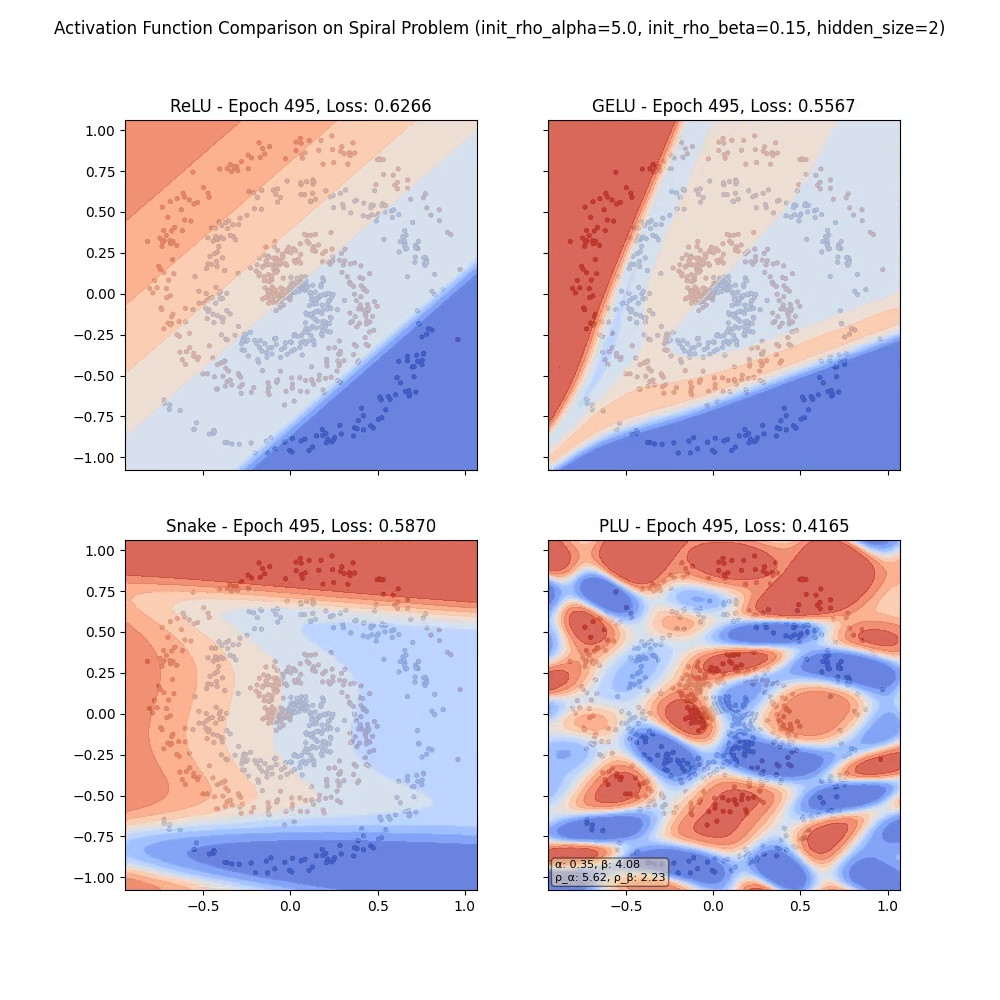}
    \caption{Final decision boundaries at Epoch 495 for a 2-neuron MLP. ReLU and GELU are architecturally incapable of solving the problem, producing simple lines and curves. Snake is beginning to form a more complex shape, but ultimately fails to fit the data, producing a "C" shape. PLU, however, learns a complex, multi-banded decision boundary, achieving a significantly lower loss.}
    \label{fig:spiral_2_neurons_final}
\end{figure}

The results in Figure \ref{fig:spiral_2_neurons_final} demonstrate the ReLU, GELU, and Snake models fail, producing simple linear or single-curved boundaries with high final losses (0.6266, 0.5567, 0.5870 respectively).

The PLU network, in stark contrast, achieves a final loss of \textbf{0.4165}. Remarkably, the network achieves this with only two neurons and a single, shared activation function shape.

On a separate run, it is able to achieve an even lower loss of 0.3046, the example image of which is provided in the accompanying examples in the code repository.

Perhaps more importantly however is its decision boundary: Its decision boundary is not a simple shape but a complex interference pattern. The visual result in Figure \ref{fig:spiral_2_neurons_final} for PLU resembles a marbled texture or a topographical map, with multiple, distinct classification bands.

\begin{figure}[H]
    \centering
    \includegraphics[width=0.8\textwidth]{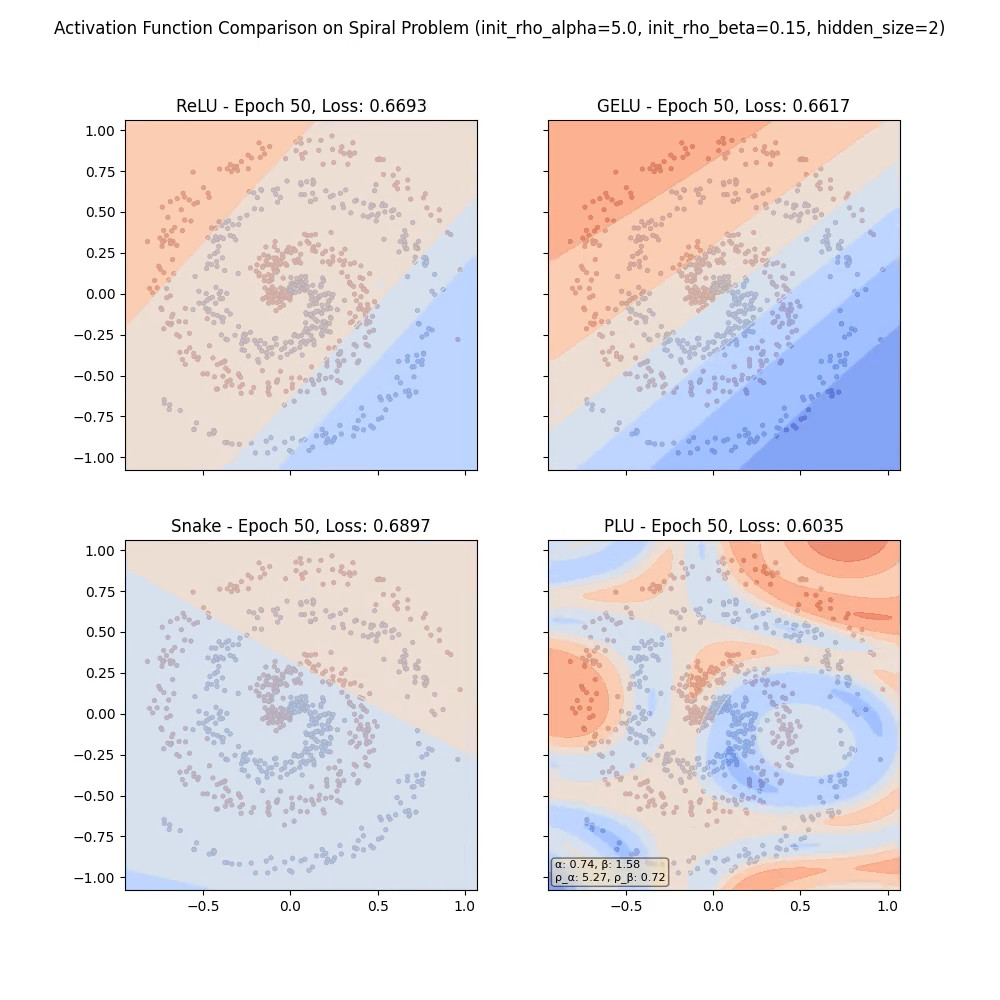}
    \caption{Decision boundaries at Epoch 50 for the same run with the 2-neuron MLP. ReLU, GELU, and Snake all consistently show semi-linear decision boundaries at this early stage. In contrast, PLU already shows a "marbled" texture, resembling a height map.}
    \label{fig:spiral_2_neurons_early}
\end{figure}

This is not specific to the end result either. As seen in Figure \ref{fig:spiral_2_neurons_early}, even at Epoch 50 in the 2-neuron MLP example, this difference in the characteristics of the decision boundaries is apparent.

\section{Discussion: From Taylor Series to Fourier Synthesis}

The performance of the Periodic Linear Unit, particularly in the 2-neuron experiment, is not merely an incremental improvement. It points to a fundamental paradigm shift in how we can conceptualize the function of a neural network.

\subsection{The Standard Paradigm: Intelligence in the Connections}

A traditional Multi-Layer Perceptron (MLP) using simple, generally monotonically-increasing activations like ReLU can be understood as a \textbf{piecewise linear function approximator.} As described by the Universal Approximation Theorem, these networks build complex functions by combining a vast number of simple "hinges" or linear pieces. This is conceptually analogous to a \textbf{Taylor series expansion}, where a function is approximated locally by a series of polynomials.

In this paradigm, the intelligence of the network resides almost entirely in the \textbf{connections}, the weights of the linear layers. The neurons themselves are simple, fixed switches. To model a complex, non-linear function like a spiral, an enormous number of these simple neurons must be precisely coordinated by an even larger number of weights. This explains the need for large, massively-parameterized models.

\subsection{The PLU Paradigm: Intelligence in the Neuron}

The PLU-based network operates on a completely different principle. It functions as a \textbf{Fourier-like synthesizer}.

Consider a standard MLP. The output of a neuron is $a(w \cdot x + b)$, where $a$ is the activation. For a network with one hidden layer, the final output is a sum of these terms: $f(x) = \sum_j v_j a(w_j \cdot x + b_j)$.

With ReLU, $a(z) = \max(0, z)$, the network becomes a sum of weighted "hinges," forming a piecewise linear approximation akin to a Taylor series. This is not limited to ReLU, either, as in general, any monotonically-increasing activation (or those close to it for much of its domain) ends up not changing the gradient landscape fundamentally. It is a "transparent" step through which gradients that are positive remain positive, and gradients that are negative remain negative.

With PLU however, the activation is $\text{PLU}(z) \approx z + \beta_{\text{eff}} \sin(\alpha_{\text{eff}} z)$, and by substituting PLU into the standard perceptron formula, we see what is really happening.
\begin{equation}
    f(x) = \sum_j v_j \left( w_j \cdot x + b_j + \beta_{\text{eff}} \sin(\lvert\alpha_{\text{eff}}\rvert (w_j \cdot x + b_j)) \right)
    \label{eq:fourier_synthesis}
\end{equation}
The output of a single neuron is now a learnable basis function, a sine wave superimposed on a line, and a shallow single-layer PLU network now becomes a \textbf{superposition of these learnable waves}.

The weights $w_j$ and bias $b_j$ control the frequency and phase of each wave, while the learnable activation parameters $\alpha$ and $\beta$ tune the shape of the basis function itself. The outer weights $v_j$ determine the amplitudes of each component wave. Fundamentally, the network is therefore not approximating with lines, but instead performing a \textbf{Fourier-like synthesis}, where the final decision boundary is the \textbf{constructive and destructive interference pattern} of the underlying waves. This is vividly illustrated in the 2-neuron experiment (Figure \ref{fig:spiral_2_neurons_final}), where the "marbled" texture is a direct visualization of wave interference.

The foundational mathematical principle of Fourier analysis states that any complex, periodic signal can be decomposed into a sum of simple sine and cosine waves. Our experiments demonstrate that a PLU-based network leverages this principle to approximate functions with staggering efficiency.

A network with ReLU activations learns a function by adding together thousands of tiny lines. A network with PLU learns a function by adding together a few powerful, adaptable waves. Because a sine wave is an infinitely more expressive basis function for describing periodic or curved phenomena than a straight line, the PLU-based network requires exponentially fewer parameters to achieve a superior result.

\subsection{Deeper PLU Networks}

It is then important to acknowledge how the dynamics change further once PLU layers are applied in a deep network, like the 2-2-2-1 model from our experiment. When there are multiple PLU layers, the operation is no longer just a simple sum, but a composition of functions, leading to nested structures akin to $\sin(\dots\sin(x)\dots)$ since the $w_j \cdot x + b_j$ term itself is a wave past the very first layer. This then brings the network beyond even simple Fourier analysis, which is additive, into a more complex, "higher-degree" sinusoidal synthesis.

While hard to conceptualize, this process is conceptually analogous to Frequency Modulation (FM) synthesis. In FM synthesis, the output of one wave is used to modulate the frequency of another, creating far richer and more complex resultant waves than simply adding them together.

A deep PLU network can be thought of as a super-charged version where, at each subsequent layer beyond the first, the input to a neuron's PLU activation ($w_j \cdot x + b_j$) becomes itself a complex wave derived from the superpositions of every single one of the generated waves from the previous layer. This input waveform is then passed through its own PLU activation which is itself a waveform modulation. The complexity and power packed into a single neuron in this scenario is immense, and yet beautifully, it is derived entirely from just a change in the activation function.

\subsection{A Paradigm Shift}

This work demonstrates that by enriching the neuron itself, by moving intelligence from the connections into the activation function, we can create models that are not only more parameter-efficient but also learn via a fundamentally different and more direct mechanism.

The success of the 2-neuron PLU model is not just a curiosity; it is proof that two well-chosen, adaptable basis functions can be more powerful than hundreds of simple ones. This challenges the long-held view of the neuron as a simple non-linearity and reframes it as a tunable, expressive function approximator in its own right.

\section{Conclusion}

We have presented the Periodic Linear Unit (PLU), a novel activation function featuring a novel formulation and a paired "repulsive reparameterization" technique that ensures training stability while enabling immense expressive power. The results are definitive, that an MLP with just two PLU neurons can solve the spiral classification problem, a task that is architecturally impossible for its counterparts using standard activations. This exponential gain in parameter efficiency stems from a fundamental paradigm shift. Instead of approximating functions with thousands of simple linear pieces as in a Taylor series, the PLU network acts as a Fourier synthesizer, learning to combine a few powerful, adaptable wave-like basis functions. This work demonstrates that placing more "intelligence in the neuron" via more expressive activations is a promising and highly efficient path toward building more powerful neural networks.

\section*{Code Availability}

The code implementing PLU and the experiments presented in this paper are both publicly available at \url{http://github.com/bill13579/plu_activation}.

\bibliographystyle{plain}
\bibliography{references}

\appendix
\section{The Benefit of Odd Activations in the Context of Signal Modelling}

A primary benefit this activation function has that was put into consideration when using it within the context of audio, which was where the idea was originally formulated, was the fact that it is odd, as mentioned earlier. This allows for the model to easily learn destructive cancellation, unlike with somewhat even functions and other asymmetric activations, where that would require complex scaling to be learned by the model itself. In particular, with parallel convolutions that are summed, this sort of constructive/destructive interference is a powerful asset as we are modelling audio. We can achieve a similar effect in that situation by just having the activation after the convolutions are summed, but that introduces another inefficiency.
\newline\newline
\textbf{Previous Approach:} $\text{Activation}(\text{Conv}_A(x) + \text{Conv}_B(x))$
\newline\newline
\textbf{Suggested:} $\text{OddActivation}(\text{Conv}_A(x)) + \text{OddActivation}(\text{Conv}_B(x))$
\newline\newline
\textit{An odd activation will allow constructive/destructive interference here, while simultaneously maintaining non-linearity for each convolution individually, thus creating two powerful non-linear functions for the price of one.}
\newline\newline
A neural network must squeeze as much "representational power" as possible out of every operation. Convolutions are linear operations, and summing multiple linear operations still gives you a linear operation. In gradient descent in general, non-linearities are the cornerstones which allows a complex linear transform to actually gain deep representational power, and by doing the activation after, we end up throwing away that power we could have gained, smearing the multi-scale convolutions together. This is not ideal. Thus, we instead use a periodic activation on the convolution outputs before summing, allowing for interference, but retaining the representational power of the different convolution "heads". The periodicity then provides both an ingrained periodic structure, as well as a non-linearity.

\end{document}